\documentclass[authorversion,nonacm, %
     sigconf = true,    % Style for GECCO proceedings (two columns and such).
     screen = true,     % Highlights links.
     review = false,     % Shows line numbers.
     anonymous = false,  % Anonymize author information.
]{acmart}

\usepackage{booktabs,subcaption,amsfonts,dcolumn}
\AtBeginDocument{%
  \providecommand\BibTeX{{%
    \normalfont B\kern-0.5em{\scshape i\kern-0.25em b}\kern-0.8em\TeX}}}

\copyrightyear{2022} 
\acmYear{2022} 
\setcopyright{acmlicensed}\acmConference[GECCO '22]{Genetic and Evolutionary Computation Conference}{July 9--13, 2022}{Boston, MA, USA}
\acmBooktitle{Genetic and Evolutionary Computation Conference (GECCO '22), July 9--13, 2022, Boston, MA, USA}
\acmPrice{15.00}
\acmDOI{10.1145/3512290.3528832}
\acmISBN{978-1-4503-9237-2/22/07}

\usepackage{xcolor,colortbl}
\usepackage{caption}
\usepackage{subcaption}
\definecolor{grey}{rgb}{0.1,0.1,0.1}

\begin{document}

\title{The Importance of Landscape Features for Performance Prediction of Modular CMA-ES Variants}

% Exploratory Analysis of the Importance of the Landscape Features for performance prediction of modular CMA-ES variants
% Linking Feature Importance to Algorithm Modules for Personalized Supervised Learning
%via Explainable Exploratory Analysis
% Analyzing the Impact of Landscape Features for 
% Explainable Exploratory Analysis for Understanding the Impact of the Landscape Features on the Individual CMA-ES Module Performance

\author{Ana Kostovska}
\orcid{0000-0002-5983-7169}
\affiliation{%
  \institution{Jo\v{z}ef Stefan Institute \&}
  \institution{Jo\v{z}ef Stefan International Postgraduate School}
  \streetaddress{Jamova cesta 39}
  \city{Ljubljana}
   \country{Slovenia}
  \postcode{1000 }
}

\author{Diederick Vermetten}
\orcid{0000-0003-3040-7162}
\affiliation{
  \institution{Leiden Institute for Advanced Computer Science}
  \city{Leiden}
  \country{The Netherlands}
}

\author{Sa\v{s}o D\v{z}eroski}
\orcid{0000-0003-2363-712X}
\affiliation{%
  \institution{Jo\v{z}ef Stefan Institute \&} 
  \institution{Jo\v{z}ef Stefan International Postgraduate School}
  \streetaddress{Jamova cesta 39}
  \city{Ljubljana}
  \country{Slovenia}
  \postcode{1000 }
}

\author{Carola Doerr}
\orcid{0000-0002-5983-7169}
\affiliation{%
  \institution{Sorbonne Université, CNRS, LIP}
  \city{Paris}
   \country{France}
  \postcode{1000 }
}

\author{Peter Korosec}
\orcid{0000-0003-4492-4603}
\affiliation{%
  \institution{Jo\v{z}ef Stefan Institute}
  \streetaddress{Jamova cesta 39}
  \city{Ljubljana}
   \country{Slovenia}
  \postcode{1000 }
}

\author{Tome Eftimov}
\orcid{0000-0001-7330-1902}
\affiliation{%
  \institution{Jo\v{z}ef Stefan Institute}
  \streetaddress{Jamova cesta 39}
  \city{Ljubljana}
   \country{Slovenia}
  \postcode{1000 }
}


\begin{abstract}
Selecting the most suitable algorithm and determining its hyperparameters for a given optimization problem is a challenging task. Accurately predicting how well a certain algorithm could solve the problem is hence desirable. Recent studies in single-objective numerical optimization show that supervised machine learning methods can predict algorithm performance using landscape features extracted from the problem instances. 

Existing approaches typically treat the algorithms as black-boxes, without consideration of their characteristics. To investigate in this work if a selection of landscape features that depends on algorithms' properties could further improve regression accuracy, we regard the modular CMA-ES framework and estimate how much each landscape feature contributes to the best algorithm performance regression models. Exploratory data analysis performed on this data indicate that the set of most relevant features does not depend on the configuration of individual modules, but the influence that these features have on regression accuracy does. In addition, we have shown that by using classifiers that take the features’ relevance on the model accuracy, we are able to predict the status of individual modules in the CMA-ES configurations.

\end{abstract}

%   <concept>
%       <concept_id>10010147.10010257</concept_id>
%       <concept_desc>Computing methodologies~Machine learning</concept_desc>
%       <concept_significance>500</concept_significance>
%       </concept>
%   <concept>
%       <concept_id>10010147.10010257.10010293.10010319</concept_id>
%       <concept_desc>Computing methodologies~Learning latent representations</concept_desc>
%       <concept_significance>500</concept_significance>
%       </concept>
%   <concept>
%       <concept_id>10010147.10010257.10010258.10010259</concept_id>
%       <concept_desc>Computing methodologies~Supervised learning</concept_desc>
%       <concept_significance>500</concept_significance>
%       </concept>
%   <concept>
%       <concept_id>10003752.10003809</concept_id>
%       <concept_desc>Theory of computation~Design and analysis of algorithms</concept_desc>
%       <concept_significance>500</concept_significance>
%       </concept>
%  </ccs2012>

% \ccsdesc[500]{Computing methodologies~Machine learning}
% \ccsdesc[500]{Computing methodologies~Learning latent representations}
% \ccsdesc[500]{Computing methodologies~Supervised learning}
% \ccsdesc[500]{Theory of computation~Design and analysis of algorithms}


\maketitle

\section{Introduction}

With the growth of the field of optimization, many algorithms have been built upon again and again, leading to vast families of algorithms that share a core design aspect. While often-times, these variants are introduced on their own, combining their contributions can allow for the creation of many more novel algorithm configurations. In particular, this can be achieved by modularizing these contributions and allowing these modules to be swapped out at will, essentially creating a configurable algorithm design space based upon a specific core algorithm, such as Genetic Algorithms~\cite{Amine21,modular_GA}, Simulated Annealing~\cite{modSA}, Particle Swarm Optimization (PSO)~\cite{PSOX}, and hybrids between Differential Evolution and PSO~\cite{boks_PSO_DE}. 

In this work, we focus on the family of Covariance Matrix Adaptation Evolution Strategies (CMA-ES)~\cite{hansen_adapting_1996}. We make use of the ModCMA~\cite{van_rijn_evolving_2016, nobel_modcma_assessing}, which implements 11 modules that can be combined arbitrarily to create thousands of variants of CMA-ES. The modules available range from elitism in the selection procedure, mirrored and orthogonal sampling mechanisms~\cite{wang_mirrored_2014, auger_mirrored_2011} to local restart strategies with changing population sizes~\cite{auger_restart_2005, bipop}. Even though all of these methods have been proposed in isolation, it has been shown that combinations of these modules can lead to configurations of the modular CMA-ES which significantly outperform all of the base variants~\cite{van_rijn_algorithm_2017}. 

Several studies explored the family of modular CMA-ES in different learning settings including exploratory analysis of CMA-ES~\cite{de2021explorative}, automated algorithm performance prediction~\cite{trajanov2021explainable}, automated algorithm selection~\cite{jankovic2020landscape}, and automated algorithm configuration~\cite{prager2020per,BelkhirDSS17}.

De Nobel et al.~\cite{de2021explorative} analyzed the CMA-ES behavior by using time-series features extracted from its dynamic strategy parameters. Their results showed that these features can be used to classify isolated CMA-ES modules and have the potential to be used in the prediction of their performance. 

Trajnov et al.~\cite{trajanov2021explainable} learned an explainable automated performance prediction model by using the landscape features of the problem instances~\cite{mersmann2011exploratory} to predict the quality of the solution after some fixed budget (i.e., number of function evaluations). Independent models were trained for different modular CMA-ES configurations and each configuration was further represented using the Shapley values that provide explanations on global (i.e., all problem instances that were involved) and local level (i.e., for each problem instance separately). Such representations allow them to distinguish between different modular CMA-ES configurations.

Jankovic and Doerr~\cite{jankovic2020landscape} used the modular CMA-ES to perform automated algorithm selection in a fixed-budget setting. By using the landscape features of the problem instances they learned performance regression models that were further used to select the appropriate modular CMA-ES configuration.

All aforementioned studies are only a part of a wide range of studies that contribute to understanding the behavior of modular CMA-ES~\cite{van2018towards,hansen2019global,vermetten2019online,vermetten2020integrated}. Even though great effort has been put in this research direction, most of the studies in automated algorithm performance prediction and selection are treating the CMA-ES configuration as a black-box system. % consists of different modules. This was done in the previously described studies~\cite{trajanov2021explainable,jankovic2020landscape}, 
That is, they do not aim at exploring the impact that each module has on the final performance of the CMA-ES configuration. In ~\cite{de2021explorative}, the time-series features have been used to classify only the isolated CMA-ES modules, but there is no information on how these features are linked when those modules are combined to create a new CMA-ES configuration. 
Prager et al.~\cite{prager2020per} investigated a problem instance-based configuration model that selects the optimal CMA-ES modules to create a configuration specific to a given problem instance. For this purpose, they trained random forest classifiers into a classifier chain scenario that uses the landscape features of the problem instance. The model is trained in a supervised learning setting and can predict if some modules should be activated or not. This study allows a selection of CMA-ES modules utilizing the landscape features of the problem instances, however without explanation of which landscape features are important for each module separately.

\textbf{Our contribution:} In this paper, we propose an approach for exploring which landscape features calculated for the problem instances are contributing to the performance prediction of the CMA-ES modules. Compared with the work done in~\cite{prager2020per}, where the activation of a module is predicted from the landscape features, in our approach we take the opposite direction, i.e., we measure how much each feature contributes to the accuracy of the best regression models. We then perform an exploratory data analysis to investigate whether different features are needed to accurately predict the performance of the modular CMA-ES variants, and if these features can be linked to individual modules. Interestingly, we find that the \emph{set} of features that are most relevant for an accurate regression does \emph{not} depend on the configuration of the modules, indicating that feature selection does not necessarily need to be personalized to the specific configuration. The \emph{relevance} of each feature, in contrast, can indeed be shown to depend on the characteristics of the algorithms. We show this by training a classifier that takes as input the vector that represents features' influence on the model accuracy (measured by approximated Shapley values in our analysis) and that outputs the status of individual modules of the CMA-ES configuration. For example, it is possible to predict whether or not a feature relevance vector stems from a configuration with \emph{elitism} turned on or off. Interestingly, this can be predicted with good accuracy even when aggregating the feature relevance vectors across different budgets that the regression models were trained for, indicating that the impact of the landscape features is relatively stable for different stages of the optimization process.

\textbf{Outline:} %or: Organization of the paper
% The remainder of the paper is organized as follows: 
Sec.~\ref{sec:background} provides relevant background for our work. Sec.~\ref{sec:methodolgy} presents the pipeline for our exploratory analysis. Key results are presented in Sec.~\ref{sec:results}, followed by a discussion of of our main findings and relevant steps for future work in Sec.~\ref{sec:conclusion}.

% TODO: Move this to a citation?
\textbf{Reproducibility:} 
% Landscape data is taken from \tome{Ana link to zenodo ?}. 
For our analysis, we trained a total number of 324 regression models for each of the 40 selected modular CMA-ES variants. We investigated fixed-budget performance for five different budgets and we considered two dimensions, $D=5$ and $D=30$. All performance data is available at~\cite{Zenodo_repo}. Landscape data, code, and our results are available at~\cite{github-repo}. 

\section{Background}
\label{sec:background}
In this section, we present the background of the two components that are parts of our proposed pipeline. 
We start by explaining exploratory landscape analysis (ELA), which is an approach for calculating numerical representations of the problem instances via features that describe their characteristics. In Sec.~\ref{sec:21}, we will then explain the process of learning a supervised machine learning model (i.e., an ELA-based performance regression model) that can be used to predict the quality of a solution achieved by the algorithm in a fixed-budget scenario.
\subsection{ELA}
\label{ssec:ela}
The problem space covers optimization problems that can belong to the same or to different problem classes~\cite{bartz2020benchmarking}. A problem class can have many different problem instances. The differences are a result of transformation processes such as translation, shifting, and/or scaling of the problem instance in the given problem space. These transformations can be stochastic in nature, defined as random variables, where input vectors are rotated, shifted, and/or scaled by predefined matrices and vectors. As a result, different problem instances of the same problem class can have different characteristics. 
To describe these properties of the problem instances, landscape analysis is applied. It calculates features by applying mathematical and statistical methods, which include different sampling methods and sample sizes. For example, exploratory landscape analysis (ELA), first introduced in~\cite{mersmann_exploratory_2011}, was designed to support the design of black-box optimization algorithms through a set of ML-based recommendations to obtain algorithms that best fit the problem at hand. The main objective of ELA is to capture the characteristics of optimization problems with a set of features, referred to as ELA features. These features are then used as input to the ML pipelines that produce the recommendations that guide the process of algorithm design. As we are dealing with black-box optimization, these features have to be calculated from an (ideally small) set of samples of the problem instance. The ELA features can be computed using the R-package \emph{flacco}~\cite{kerschke_r-package_2016}. It contains 343 different ELA features grouped into 17 feature sets~\cite{kerschke_automated_2019}.

\subsection{ELA-based performance regression model}\label{sec:21}

% To train an ELA-based performance regression model, first we should involve a set of benchmark problem instances for which the performance of the algorithm is known . Next, using the ELA approach, we should calculate the numerical representation of the problem instances.  The idea behind training a regression model is to link the ELA representation of the problem instance with the performance of the algorithm achieved on that problem instance. The performance of the algorithm can be investigated in a fixed-target (i.e., returns the budget required to achieve a certain target) or a fixed-budget scenario (i.e., estimates the quality of the solution achieved with a given computational budget).

% In this paper we focus on ELA-based performance prediction in a fixed-budget scenario (i.e., after some function evaluations have been performed).

The core motivation of creating ELA-based performance regression models lies in their ability to link the landscape features of the problem instance with the performance achieved by the algorithm on that problem instance. To collect the necessary training data for these models, we need to make use of a set of benchmark problem instances for which we have gathered performance data of our algorithms. This performance data can be viewed from two main perspectives: fixed-target (i.e., returns the budget required to achieve a certain target) and fixed-budget (i.e., estimates the quality of the solution achieved with a given computational budget). This performance data can then be linked to the ELA features collected on the specified benchmark function instances to train the regression models.

In this paper, we focus on the fixed-budget approach, where we predict the mean function value of the best solution that is reached by the algorithms after $X$ function evaluations.

\section{Exploratory analysis pipeline}
\label{sec:methodolgy}
\begin{figure*}
    \centering
    \includegraphics[width=\linewidth]{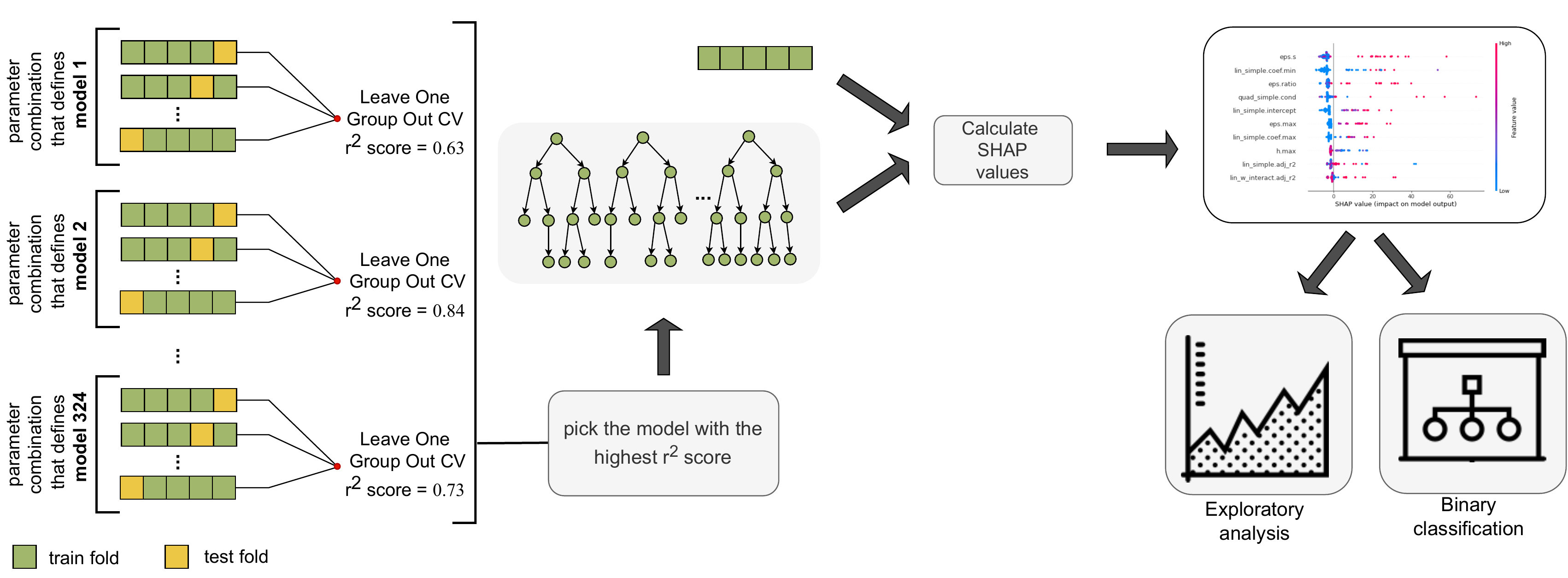}
    \caption{Overview of the exploratory analysis pipeline for linking problem landscape data with the performance of isolated modules of the CMA-ES algorithm.}
    \label{fig:pattern}
\end{figure*}
The proposed exploratory analysis pipeline consists of three main parts, i.e., learning a model for automated prediction of the algorithm's performance, feature importance calculation and ranking, which are further used to perform exploratory analysis and training classifiers to predict the status of a given CMA-ES module. The flowchart of the pipeline is presented in Fig.~\ref{fig:pattern}. 

\noindent\textbf{Regression model and its hyperparameters:} To learn an algorithm performance prediction model, we took a supervised approach where each problem instance is described with a vector of $n$ ELA features $(x_1, x_2,...,x_n)$. These vectors are associated with one real value $y$ indicating the performance of the algorithm on a specific problem instance after a fixed budget. Since the target that is being predicted is a real value, we are dealing with a regression task. Many studies have already tried to investigate this task in the context of automated algorithm performance prediction~\cite{munoz2012meta}. Here, we are learning Random Forest (RF) regression models where we also considered tuning part of the hyperparameters. Previous studies have already shown that RF can provide promising results dealing with automated performance prediction~\cite{jankovic2021impact}. For hyperparameter tuning, we employed the grid search methodology, which performs an exhaustive search over a manually selected finite subset of candidate solutions of the hyperparameter space of the algorithm. 

\noindent\textbf{Leave-One-Group-Out validation and selection of the best model:} For validation of the induced models, we use the Leave-One-Group-Out strategy. Let us assume that for each problem, $m$ instances are involved. In our case, groups are denoted by the problem instances, which means that we ended up with $m$ groups.
%We consider the first 5 instances of each of the 24 BBOB functions, which leaves us with 5 different groups. \dv{It seems a bit odd to me that BBOB is discussed here already, but only introduced in 4.1}
% This approach ensures that one problem instance does not appear in both the training and the testing set at the same time and avoids reporting overly optimistic results.
The evaluation metric of choice is the $r^2$ score. The same evaluation metric is used for evaluating the individual model candidates and as a heuristic in the grid search step, where based on the $r^2$ score we identify the best performing model (the model with the highest $r^2$ score). 

\noindent\textbf{Calculating the explanations of ML regression model:} Once the predictive regression model is learned, we then proceed with the calculation of the Shapley values, i.e. the ELA feature importance scores. The Shapely values are quantification of the marginal contribution of each input feature on the predictions made by the predictive model \cite{molnar2020interpretable}. The computation of the Shapley values is computationally expensive as it considers each possible combination of features (a power-set of features) to determine their importance on the prediction. We therefore apply the SHAP (SHaply Additive exPlanations) algorithm~\cite{lundberg2017unified}. The SHAP algorithm was first introduced in 2017 as a game-theoretic approach to explaining the output of ML models. Since then, it has successfully been applied to reverse-engineer the output of predictive algorithms (including the so-called black-box algorithms). 

\noindent\textbf{Exploratory analysis:} The pipeline for learning a regression model and obtaining the SHAP values to explain its predictions is repeated for each algorithm configuration considered in the study (explained in Sec.~\ref{sec:perf_data}). The algorithm configurations are then grouped with respect to a given module to observe whether there are some differences in the performance when there are some specific structural changes to the algorithm. For simplicity of the analysis, the groups of algorithm configurations are set to always differ in one module of the CMA-ES algorithm. For example, in the modular implementation of the CMA-ES algorithm, the elitism module can be either turned on or off, while the step-size-adaptation module, if active, can take 2 different values, i.e., csa and psr.
% while the bound-correction module, if set, can take different values, e.g., unif\_resample, saturate, COTN, toroidal, and mirror.
Such an approach facilitates the exploratory analysis of the effect each of these models has on the final performance of the algorithm. Furthermore, trends in the landscape space can be observed. More specifically, some ELA features can be found to hold more predictive value than the rest by observing the SHAP values across the different problems in multiple dimensions and different thresholds for the maximum number of function evaluation runs. 

\noindent\textbf{Classifiers for predicting the status of a CMA-ES module:} Using the calculated Shapley values aggregated across all benchmark problem instances, we can generate a vector representation for each configuration. This representation is a vector of Shapley values, one per each ELA feature, which convey the information of how the ELA feature contributes to the prediction of the performance of the configuration. Further, these representations can be labeled with regard to the activation status of a module in modular CMA-ES (e.g., taking into account the elitism, the labels will be ``ON" and ``OFF"). Using the annotated data, we can learn classifiers that can predict from this representation  the status of a given CMA-ES module.

%\peter{Should we also mention here that SHAP values also greatly depends on ML used, but since we will focus on RF only, we will in some way omit/neglect this influence in our study?}
%Nice comment for the extension :)

\section{Results \& Discussion}
\label{sec:results}

% This section starts with the explanation of the experimental setup involving the landscape and performance data together with the hyperparameters utilized in the grid search of the RF models. Next, we are going to present the analysis where we are investigating the impact of elitism and step-size modules in modular CMA-ES performance.
This section describes the experimental setup, which includes the description of landscape and performance data, as well as the details of hyperparameter tuning used in the grid search for the RF models. Following that, we will present the analysis in which we examine the effect of elitism and step-size modules on modular CMA-ES performance.

\subsection{Experimental setup}
\subsubsection{Landscape data} 
\label{sec:landscape_data}
The problem portfolio consists of the first five instances of the 24 noiseless BBOB functions~\cite{bbob} of the COCO benchmark environment~\cite{hansen2020coco} in both $D=5$ and $D=30$ dimensions. Thus, we obtain two problem sets (one for each dimension) containing 120 problem instances each. The problem landscape data are represented using 46 of the so-called "cheap" ELA features implemented in the R package \texttt{flacco}~\cite{kerschke_r-package_2016}. 
%\peter{We "always" use 46 features, should not we also make experiments for 15 invariant features, to check/remove the influence of "invariance". This is more a comment/suggestion for other/future experiments.} We can do this in the extension.
Here, however, we do not calculate the ELA features but instead, make use of a publicly available dataset that contains the necessary landscape data~\cite{quentin_renau_2020_3886816}. The ELA features utilized in the study can be divided into 6 groups, i.e., dispersion, y-distribution, meta model, information content, nearest better clustering, and  principal component analysis. The selected ELA features have been calculated using the Sobol sampling strategy with $100D$ sample size on a total of 100 independent repetitions. For a more robust analysis, we use the median of the 100 calculated feature values. This data is also made easily accessible via the OPTION ontology~\cite{kostovska2021option} and we used their API to extract it.

\subsubsection{Performance data}
\label{sec:perf_data}
Since collecting data based on a complete enumeration of all combinations of modular CMA-ES modules is computationally infeasible, we make use of a set of 40 configurations. The details of which ones were selected and used can be found in our Zenodo repository~\cite{Zenodo_repo}. For each of these configurations, we collect 10 independent runs on each of the first 5 instances of the BBOB functions, in both 5 and 30 dimensions. For each of these runs, we then extract the best precision reached after $B=\{500,2\,000,5\,000,10\,000,50\,000\}$ function evaluations. 

\subsubsection{RF hyperparameter tuning}
For learning the RF models for automated performance prediction, we used the Python package scikit-learn~\cite{pedregosa2011scikit}. 
The hyperparameters of the RF model are estimated with the Grid Search strategy. We considered tuning 5 different RF parameters: (1) n\_estimators - the number of trees (estimators) in the forest; (2) max\_features - the number of features considered when making the best split ; (3) max\_depth - the maximum depth of the tree; (4) min\_samples\_split - the minimum number of samples required for splitting a node in the tree; and (5) criterion - the function that measures the quality of a split. The values for the aforementioned hyperparameter used in the grid search are presented in Tab.~\ref{tab:hyperparameters}. This resulted in training 324 model candidates for each modular CMA-ES algorithm configuration involved in the study.
Note that we have 40 different algorithm configurations. However, since we consider problem instances in 2 different dimensions ($5D$ and $30D$) and we generate performance data for 5 budgets, we essentially end up learning 400 ($40 \times 2 \times 5$) predictive models. 
Hence a total of $129\,600$ ($324 \times 400$) RF models have been trained for the purpose of this study. The best model for each of the 400 configuration was selected using the $r^2$ evaluation metric.
\begin{table}
\begin{center}
\caption{RF hyperparameters and their values used in the grid search.}
\label{tab:hyperparameters}
\begin{tabular}{ cc }
\toprule
 hyperparameter & search space \\ 
\midrule
 n\_estimators & $[100, 500, 1000]$\\
 max\_features & $['auto', 'sqrt', 'log2']$ \\ 
 max\_depth & $[4,8,15, None]$ \\
 min\_samples\_split & $[2, 5, 10]$ \\
 %criterion & $['squared\_error', 'absolute\_error', 'poisson']$ \\
 criterion & \begin{tabular}[x]{@{}c@{}}$['squared\_error', 'absolute\_error',$\\$'poisson']$\end{tabular} \\
 \hline
\end{tabular}
\end{center}

\end{table}

\begin{table*}[]
    \centering
    \caption{An example pair of algorithm configurations that differ only in one of the modules of the CMA-ES algorithm, i.e., in the elitism module (active and non-active).}
    \label{tab:pair_conf}
    \begin{tabular}{ccccccc}
      \toprule
      elitist & mirrored & base\_sampler & weights\_option & local\_restart & bound\_correction & step\_size\_adaptation \\
      \midrule
      \cellcolor{grey!25} True & mirrored & gaussian & default & OFF & saturate & csa \\
      \cellcolor{grey!25} False & mirrored & gaussian & default & OFF & saturate & csa \\
      \hline
    \end{tabular}
    
\end{table*}

\subsection{Impact of the elitism module}
To estimate the impact of the elitism module on the modular CMA-ES performance, we used the proposed exploratory analysis pipeline. For this purpose, from the 40 algorithm configurations, we identified 11 pairs of configurations in which all the modules are set the same, but the elitism module is either activated or not. An example of such a pair is depicted in Tab.~\ref{tab:pair_conf}. 

As mentioned previously in the exploratory analysis pipeline, for each algorithm configuration from the 11 pairs (total of 22 configurations) a separate RF with the optimal hyperparameters is learned. Following that, the SHAP algorithm was used to determine the feature importance for each of the 46 ELA features. To provide global explanations about the impact of elitism on algorithm performance, we jointly analyze the 11 pairs. 
% First, we obtain the Shapley values for all 46 ELA features for the configurations. 

\begin{figure*}[ht]
    \centering
     \includegraphics[width=0.95\linewidth]{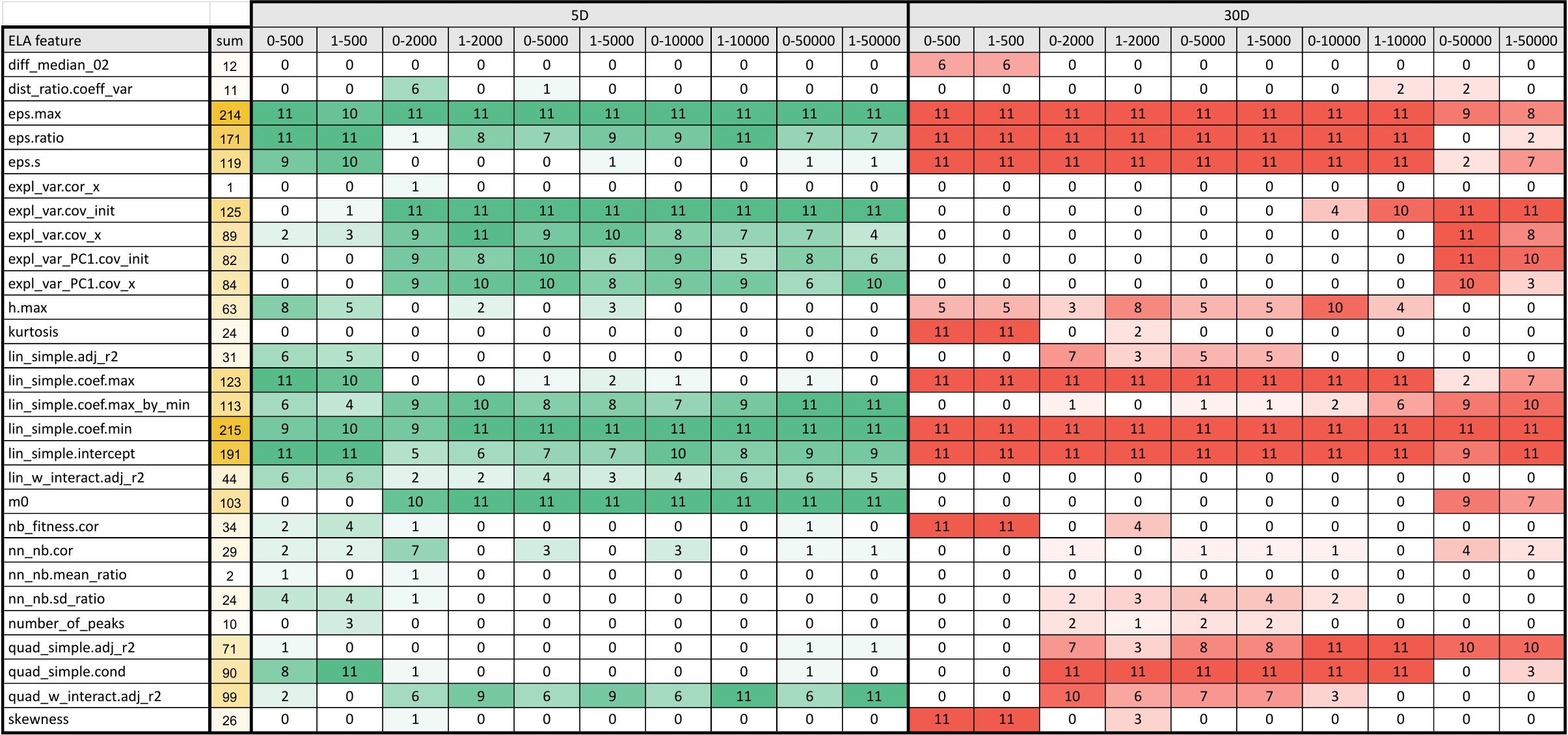}
    \caption{Frequency of appearance of the ELA features as one of the top 10 most important features in the 11 pairs of configurations of modular CMA-ES with elitism turned off and on. The results are calculated on the 24 BBOB functions in both 5 and 30 dimensions and where the best precision is reached after $B=\{500,2\,000,5\,000,10\,000,50\,000\}$ function evaluations.}
    \label{fig:elitism}
\end{figure*}

We are using the leave-one-group-out validation with 5 groups (four for training and one for testing). However, we are interested in looking at the Shapley values for the training folds only as this is the data that is used for learning the predictive models and can provide explanations on how the model works. Hence, the validation involves calculating the Shapley values four times for each benchmark problem instance (each problem instance is represented four times in the training data and once in the testing data). To obtain a single Shapley value for each ELA feature and problem instance, the mean of the four values is calculated. Additionally, to select the most important ELA features across all benchmark problem instances, we calculate the mean of the Shapley values across all problem instances. Finally, we select the top $k$ most important features to perform the exploratory analysis.

First, we have conducted an analysis using the top 10 most important ELA features. Then, we count the number of times a given feature appears in the top 10 across all pairs of problem instances in 5 and 30 dimensions and with 5 different budgets. In other words, we calculate the frequency with which a given ELA feature appears in the top 10 most important as indicated by their Shapley values. This value ranges between 0 and the total number of configuration pairs considered (in our case 11). A value of zero indicates that an ELA feature was not ranked in the top 10 in any of the 11 configuration pairs, whereas a value of 11 indicates that the feature was ranked in the top 10 in all 11 configurations. The results are shown in Fig.~\ref{fig:elitism}. Note that due to page limitation, in Fig.~\ref{fig:elitism} we only list the features that appear in the top 10 at least once.

Focusing on Fig.~\ref{fig:elitism} and $D=5$, we can see that several ELA features are unaffected by the status of the elitism module (whether the elitism module is active or not), and even appear in the top 10 most important features regardless of the budget in which the algorithm performance is predicted. These features include ``eps.max", ``lin\_simple.coef.max\_by\_min", and ``lin\_simple.coef.min". There are several features, such as ``eps.s", ``h.max",  ``lin\_simple.coev.max", and ``quad\_sim\-ple.cond" that seem to be important when predicting the performance for small budgets (500 function evaluations) both when elitism is active and not, but are not important for larger budgets. The results indicate that these features help predict the performance when the algorithm is still in the exploration phase, looking for areas in search space with a high probability of finding good solutions. The opposite is true for another group of features (``expl\_var.cor\_x", ``expl\_var.cov\_x", ``expl\_var\_PC1.cov\_init", \linebreak ``expl\_var\_PC1.cov\_x",
and ``m0"). They are not important for small budgets, however, when the budget increases ($\ge$1000 function evaluations), they become important regardless of whether the elitism module is on or off. 
%\peter{The difference when the algorithm goes from exploration to exploitation phase is not so clear cut with respect to the number of evaluations, but can be more clearly seen by the population's solutions distribution and/or convergence curve. So we must be aware of this.}
%\tome{Carola and Peter please check the following :)} This indicates that these features are important when the algorithms are in exploitation phase. \peter{To confirm this maybe we should draw convergence curve for some selected instances? If they would be in line with what we discovered in the data, this would be really great. Probably this will be much harder to show for the 30D, but at the same time maybe we will see that the algorithms are still in the exploration phase much longer... and if the data confirms this. Realllyyyy great!!!} 
The most interesting ELA feature from this analysis is ``quad\_w\_interact.adj\_r2". At the beginning of the optimization process (when the budget is set to 500 function evaluations), it is not important whether elitism is on or off, however as the budget increases, it is visible that this feature is more important in the cases when elitism is on. We can conclude that the majority of the ELA features are behaving similarly regardless of the elitism module activation state. From the results, it follows that there is no special group of features that are related to the elitism module. On the other hand, this provides promising results that the same ELA feature portfolio can be enough to predict the performance of different modular CMA-ES no matter the status of the elitism in the algorithm's configuration.

\begin{figure}
    \centering
    \begin{subfigure}[b]{\linewidth}
    \includegraphics[width=\linewidth]{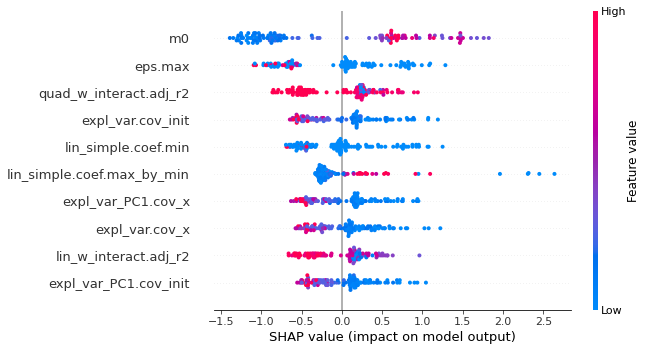}
    \caption{Elitism activated}
    \label{fig:elitism_on}
    \end{subfigure}
    \hfill
    
\begin{subfigure}[b]{\linewidth}
    \centering
    \includegraphics[width=\linewidth]{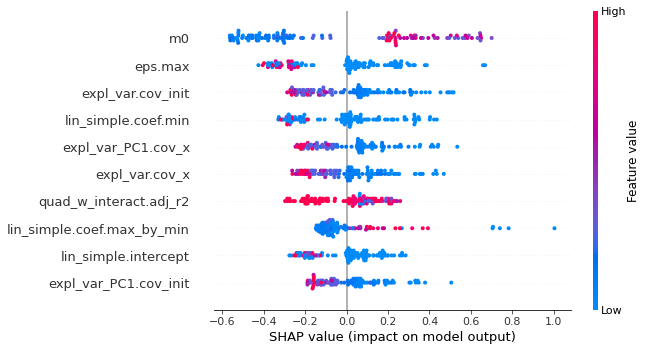}
    \caption{Elitism not activated}
    \label{fig:elitism_off}
\end{subfigure}
    
    \caption{Shapley values of the top 10 ELA features for a pair of configuration presented in Tab.~\ref{tab:pair_conf}, for $5D$ and 2000 function evaluations.}
    \label{fig:local_analysis}
\end{figure}

% After providing a global summary of the importance of the ELA features, 
After summarizing the importance of the ELA features globally, we can use the proposed exploratory analysis pipeline to provide local explanations for a pair of configurations that differ only in the elitism module. Here, we will discuss the local explanations for the pair presented in Tab.~\ref{tab:pair_conf}. Fig.~\ref{fig:local_analysis} presents the summary Shapley plots for both configurations in the pair separately. These summary plots illustrate the positive and negative relationships with the quality of solution reached after some fixed budget. Each dot in the plots represents a problem instance. The ELA features are listed in descending order of importance. The colors used indicate the magnitude of the ELA feature value (red representing higher values and blue representing lower values). Finally, the position on the horizontal axis presents the impact of the ELA feature value on the prediction of the target. We observe that regardless of whether the elitism module is activated, the top 10 most important ELA features are mostly the same, and their contributions of the prediction to the algorithm performance follow similar patterns.

\begin{sloppypar}
Focusing on $D=30$ (Fig.~\ref{fig:elitism}), similar patterns as in $D=5$ can be observed. However, we will not compare them across different dimensions since for the simplicity of analysis the selected budgets are the same and a larger budget is likely required to solve the problem instances in high dimension. Here, ``eps.max", ``eps.ratio", ``eps.s", ``lin\_simple.coef.max", ``lin\_simple.coef.min", and ``lin\_simple.coef.intercept", are important for various budgets regardless of the elitism activation status. The ``expl\_var.cov\_init", ``expl\_var.cov\_x", ``expl\_var\_PC1.cov\_init", ``expl\_var\_PC1.cov\_x", ``m0" are important only for large budgets ($50\,000$), again, regardless of the elitism activation status. The statistical features ``diff\_median\_02", ``kurtosis", and ``skewness" are important only for small budgets when the dimension is 30.
\end{sloppypar}

\begin{figure*}[ht]
    \centering
    \includegraphics[width=0.95\linewidth]{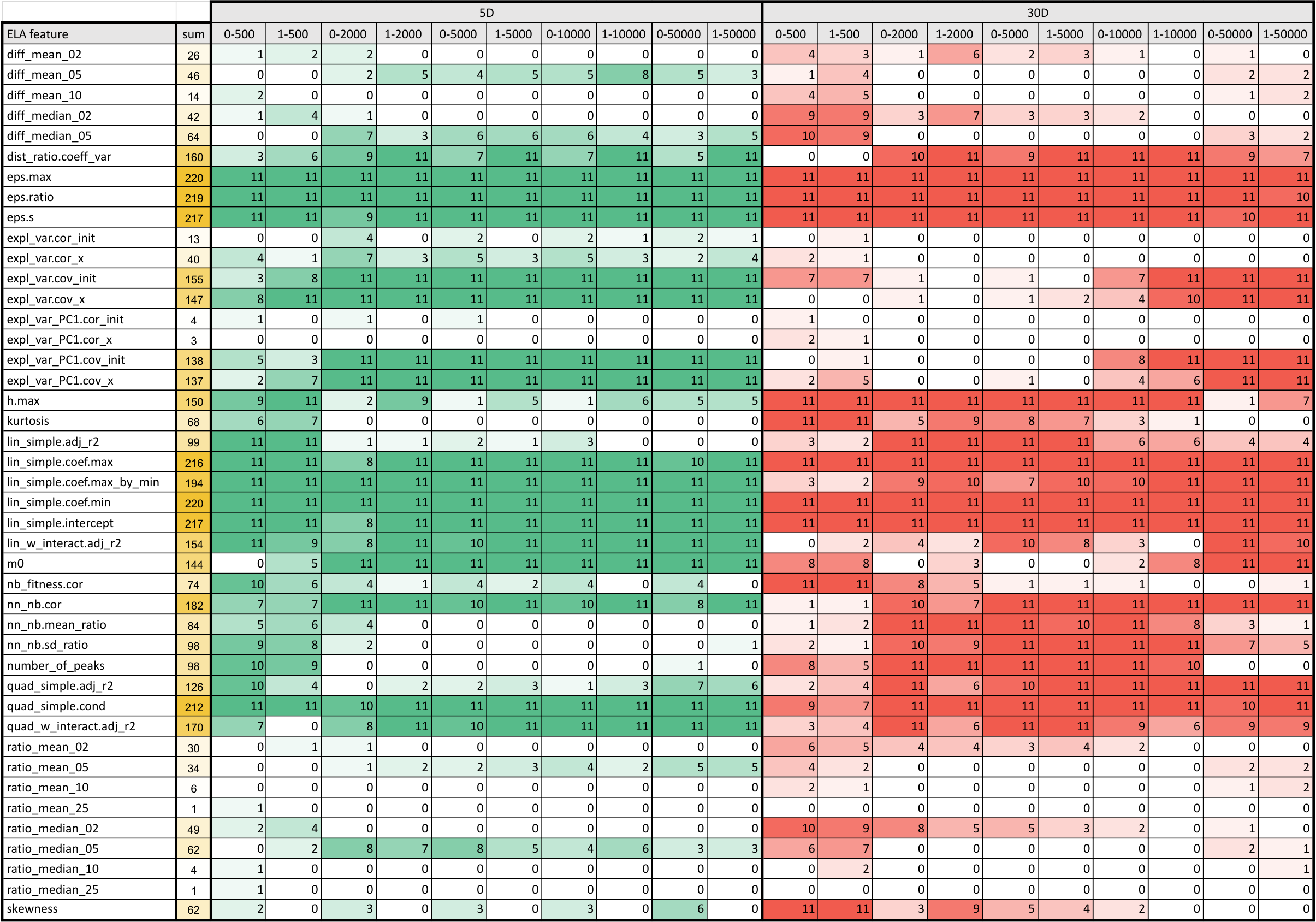}
   \caption{Frequency of appearance of the ELA features as one of the top 20 most important features in the 11 pairs of configurations of modular CMA-ES with elitism turned off and on. The results are calculated on the 24 BBOB functions in both 5 and 30 dimensions and where the best precision is reached after $B=\{500,2\,000,5\,000,10\,000,50\,000\}$ function evaluations.}
    \label{fig:top20}
\end{figure*}

To determine whether these findings hold true as the number of the most important ELA features considered increases, Fig.~\ref{fig:top20} presents the same analysis performed for $5D$ and $30D$ for all 5 budgets when the top 20 features are selected instead of the top 10. Without going into detail, we can conclude that similar patterns exist for the majority of the previously described features. The difference is that more features appear, which is logical given the selection of 20 features. Certain patterns observed in the top 10 selection do change. For instance, when focusing on $5D$, ``eps.s" and ``lin\_simple.coef.max" appear to be important also for larger budgets, which was not the case in top 10 features analysis, where they appeared only for really small budgets. It is also interesting that the feature, ``quad\_w\_interact.adj\_r2", that was more important when the elitism was on, now with the top 20 analysis seems to be equally important regardless of the activation of the elitism module.

The exploratory analysis provides us with some descriptive analysis using a counting procedure. To see if we can detect differences between the configurations when the elitism module is switched on or off, we represented each configuration as a vector of 46 Shapley values for each budget separately and then we labeled the vector with ``ON" or ``OFF" with regard to the elitism activation status. Here, we ended up with 110 (11 pairs $\times$ 2 configurations $\times$ 5 budgets) configurations (i.e., data instances) represented with 46 Shapley values that provide representations of how the ELA features are utilized to predict each configuration performance. The distribution of the labels within the 110 configurations is 50:50, so the baseline of classifying them with regard to the activation of the elitism is 50\%. The process of labeling the data was done twice, so we ended up with two datasets, one for each dimension.

Next, for each dataset separately, we trained a binary classifier that uses the Shapley representations as input data and predicts if the elitism is ON or OFF. We utilized RF for learning the classifiers and we evaluated them in 5 cross-fold validation using the 110 instances. We used RF with default parameters as implemented in the Python package scikit-learn.

Tab.~\ref{tab:class_elitism} presents the accuracy and the F$_1$ score obtained for the classification done in each problem dimension. In $5D$, the classifier provides 79\% accuracy of classifying the status of elitism (ON or OFF), while in $30D$ the accuracy is around 66\%. It is important to note that these results are obtained regardless of the budget value. In both dimensions, the accuracy is above the baseline, which shows promising results. These results indicate that an automated performance prediction model built using the explanations of training contains useful information that can be used to predict whether a configuration has the elitism module activated or not. 
% Even the fact that the importance(i.e., ranking) of the ELA features is mostly the same regardless of the activation of the elitism module using the exploratory analysis counting approach, these results point out that their contribution (aggregated Shapley values across all problem instances) is different and allow us to distinguish if a module is active or not.
\begin{figure*}[ht]
    \centering
     \includegraphics[width=0.95\linewidth]{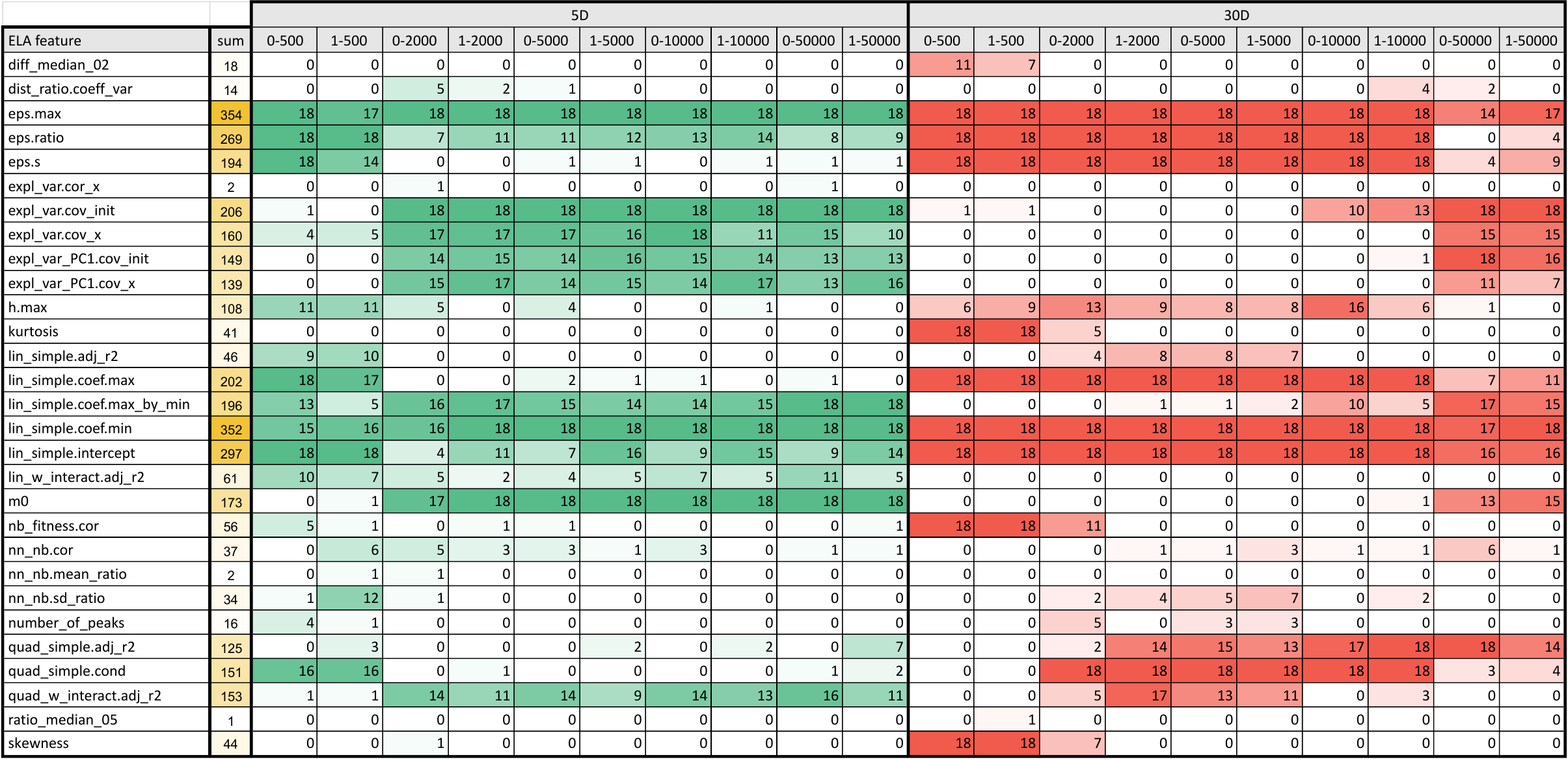}
     \caption{Frequency of appearance of the ELA features as one of the top 10 most important features in the 18 pairs of configurations of modular CMA-ES with step-size-adaptation module set to csa and psr. The results are calculated on the 24 BBOB functions in both 5 and 30 dimensions and where the best precision is reached after $B=\{500,2\,000,5\,000,10\,000,50\,000\}$ function evaluations.}
    \label{fig:step_size}
\end{figure*}
Although the importance of the ELA features is largely the same regardless of the activation status of the module, these findings show that their contribution (aggregated Shapley values across all problem instances) is different, allowing us to determine whether a module is active or not.

% \begin{table}[ht]
% \centering
% \caption{Binary classification results for the activation of the elitism module.}
% \label{tab:class_elitism}

% \end{table}

\begin{table}[h]
    \begin{subtable}[h]{0.45\textwidth}
        \centering
        \begin{tabular}{|r|c|c|}
          \hline
          & \multicolumn{2}{c|}{Problem dimension} \\
          \hline
         & $5D$ & $30D$  \\ 
          \hline
         Accuracy & 0.79090  & 0.66364\\ 
         F$_1$ score & 0.78938 &  0.66225\\ 
           \hline
        \end{tabular}
       \caption{Activation of the elitism module}
       \label{tab:class_elitism}
    \end{subtable}
    \hfill
    \begin{subtable}[h]{0.45\textwidth}
        \centering
    \begin{tabular}{|r|c|c|}
  \hline
  & \multicolumn{2}{c|}{Problem dimension} \\
  \hline
 & $5D$ & $30D$  \\ 
  \hline
 Accuracy & 0.76111  & 0.68333\\ 
 F$_1$ score & 0.75523 &  0.68149\\ 
  \hline
\end{tabular}
        \caption{Step-size-adaptation set to csa and psr}
        \label{tab:class_step}
     \end{subtable}
     \caption{Binary classification results.}
     \label{tab:classification_results}
\end{table}

% \begin{table}[ht]
% \centering
% \caption{Binary classification results for step-size-adaptation set to csa and psr.}
% \label{tab:class_step_size}
% \begin{tabular}{|r|c|c|}
%   \hline
%   & \multicolumn{2}{c|}{Problem dimension} \\
%   \hline
%  & $5D$ & $30D$  \\ 
%   \hline
%  Accuracy & 0.76111  & 0.68333\\ 
%  F$_1$ score & 0.75523 &  0.68149\\ 
%   \hline
% \end{tabular}
% \end{table}

\subsection{Impact of the step-size module}
In our second use case, we used the exploratory analysis pipeline to investigate the impact of the step-size-adaptation module that can be switched between two values: csa and psr. Here, 18 pairs of configurations that differ only in the step-size adaptation module have been identified and selected for further analysis. Fig.~\ref{fig:step_size} presents the results of the exploratory analysis performed with the selection of the top 10 most important ELA features. As a result, there appears to be no clear difference between the two possible values of this module, and the same features appear to contribute the same in both cases. Again, there are features that are equally important for all budgets, features that are important at the beginning of the optimization process but are less important for larger budgets, and features that are not important at the beginning of the optimization process but are important for larger budgets, regardless of the value of the step-size-adaptation module.

To go beyond the descriptive analysis, we have again trained classifiers for predicting the step-size adaptation (csa or psr) by using the Shapley representations of the configuration. Here, we ended up with 180 (18 pairs $\times$  2 configurations $\times$ 5 budgets) configurations (data instances) labeled with ``CSA" or ``PSR". The classifier baseline is 50\%, the same as for the elitism module. For each dimension, we learned an RF model for predicting the step-size-adaptation value. We used the same experimental setting as in the classification analysis for the elitism module (RF with default parameters and 5-fold cross-validation). The results are presented in Tab.~\ref{tab:class_step}. For both dimensions, the accuracy is above the baseline, 76\%, and 68\% for $5D$ and $30D$, respectively. These results point out that the Shapley representations of the configurations that consist of the information of how the ELA features contribute to predicting their performance are useful to predict the value of the step-size-adaption module.

\balance 

\section{Conclusions and Future Work}
\label{sec:conclusion}
In this study, we have investigated the impact of the landscape features on the performance of two modules involved in modular CMA-ES, elitism, and step-size-adaptation. The same pipeline can be applied to the other modules as well. The explanations (Shapley values for each ELA feature) from automated algorithm performance prediction are used as vector representations for each modular CMA-ES. For both investigated modules (elitism and step-size adaptation), it seems that the same ELA features are within the top 10, 15, and 20 most important features that contribute to their performance prediction (using their ranking) regardless of the possible values of each module. Hence, the main take away message is that we can do feature selection for all algorithms at once, no matter the configuration of the algorithm. However, it is advisable to train separate regression models for each configuration. 
% These results suggest that, regardless of the elitism module's activation status or the value of the step-size-adaptation module, the same feature portfolio without any feature selection will be sufficient to predict the performance of various configurations. Furthermore, by representing the configurations as a vector of real Shapley values (i.e., using their importance rather than checking if a feature appears in top $k$ features), supervised classifiers can be trained to predict whether a modular CMA-ES uses elitism or what the value of the step-size-adaptation module is, when activated.

%  if not done already, please highlight that the results are good news in that we can do feature selection once and for all algos but that it seems advisable to train individual regression models for each algo

The study has been conducted on a limited number of modular CMA-ES configurations. With more training data we would expect the classifiers' performance to improve even more. The presented pipeline can directly be used to analyze other CMA-ES modules not considered in this work.% which were not analyzed in this study due to the page limit.

Such supervised classifiers can further be used to predict the modules for which a random modular CMA-ES is composed. 
%\peter{We can write this, but I doubt it is interesting, since we might predict what could be used to achieve the results but not what actually was (it is related to my last comment below). $->$}Even more, there is a lot of benchmark performance data that is publicly available for CMA-ES. Sometimes it happens that the names of the algorithms are the same but there is a difference in the modules of which they are composed. Having the performance data, the approach can be used to detect which modules are utilized. 
Moreover, having performance data from some other algorithms such as DE or PSO obtained on the same problem instances will allow us to run the pipeline and train an explainable regression model, whose explanations will further be used as their Shapley representations. 
%\peter{I was also thinking about this idea, just do not know if it is feasible. So I would put this in future work as that we would like to test if this is possible, since this would be really interesting if it would work. $->$}
%Thanks Peter :D
Finally by using classifiers that can predict the status of each module in the modular CMA-ES, we can use the learned representations for DE or PSO with each classifier, to be able to find which modules should be activated and with which values in order to find similar modular CMA-ES configuration (to mimic the behavior of the other algorithms through the modular CMA-ES). This is a direction we plan to explore in our future work.

\begin{acks}
The authors acknowledge the support of the Slovenian Research Agency through research core grants No. P2-0103 and P2-0098, project grants No. J2-9230 and N2-0239, and the young researcher grant No. PR-09773 to AK, as well as the EC through grant No. 952215 (TAILOR). Our work is also supported by Paris Ile-de-France region, via the DIM RFSI AlgoSelect project and via a \href{http://species-society.org/scholarships-2022/}{SPECIES scholarship} for Ana Kostovska. 
\end{acks}

\bibliographystyle{ACM-Reference-Format}
\bibliography{references}

% \appendix
\end{document}